\documentclass{article}

\usepackage{arxiv}

\usepackage[utf8]{inputenc} 
\usepackage[T1]{fontenc}    
\usepackage{hyperref}       
\usepackage{url}            
\usepackage{booktabs}       
\usepackage{amsfonts}       
\usepackage{nicefrac}       
\usepackage{microtype}      
\usepackage{lipsum}		
\usepackage{graphicx}
\usepackage[numbers]{natbib}
\usepackage{doi}
\usepackage{subfigure}

\usepackage{algorithm}
\usepackage{algorithmic}
\usepackage{tikz}
\usetikzlibrary{shapes.geometric, arrows, positioning}
\tikzstyle{startstop} = [rectangle, rounded corners, minimum width=3cm, minimum height=1cm, text centered, draw=black, fill=red!30]
\tikzstyle{process} = [rectangle, minimum width=3cm, minimum height=1cm, text centered, draw=black, fill=blue!30]
\tikzstyle{arrow} = [thick,->,>=stealth]
\tikzstyle{label} = [font=\small, align=center]
\usepackage{pifont}   

\newcommand{\cmark}{\ding{51}} 
\newcommand{\xmark}{\ding{55}} 
\usepackage{amssymb}
\usepackage{mathtools}
\usepackage{amsthm}

\usepackage{pgfplots}
\usepackage{xfrac}
\usepackage{enumitem}

\usepackage{commath}
\usepackage{amsmath}
\usepackage{pgfplots}
\usepackage{xfrac}
\usepackage{amsfonts}
\usepackage{enumitem}
\usepackage{physics}
\usepackage{todonotes}
\usepackage{adjustbox}
\usepackage{multirow}
\usepackage{makecell}
\usepackage{filecontents}
\usepackage{caption}

\newcommand{\REFLOW}{\texttt{\textsc{REFLOW}}}
\usepackage[capitalize,noabbrev]{cleveref}

\usepgfplotslibrary{fillbetween}
\usepgfplotslibrary{groupplots}

\title{Signal Collapse in One-Shot Pruning: When Sparse Models Fail to Distinguish Neural Representations}

\author{
Dhananjay Saikumar \\
School of Computer Science\\
University of St Andrews\\
St Andrews, UK, KY16 9SX\\
\texttt{ds304@st-andrews.ac.uk}
\And
Blesson Varghese \\
School of Computer Science\\
University of St Andrews\\
St Andrews, UK, KY16 9SX\\
\texttt{blesson@st-andrews.ac.uk}
}

\hypersetup{
pdftitle={A template for the arxiv style},
pdfsubject={q-bio.NC, q-bio.QM},
pdfauthor={David S.~Hippocampus, Elias D.~Striatum},
pdfkeywords={First keyword, Second keyword, More},
}

\begin{document}
\maketitle
\thispagestyle{empty}  
\pagestyle{plain}      
\begin{abstract}
Neural network pruning is essential for reducing model complexity to enable deployment on resource-constrained hardware. While performance loss of pruned networks is often attributed to the removal of critical parameters, we identify \textbf{signal collapse}—a reduction in activation variance across layers—as the root cause. Existing one-shot pruning methods focus on weight selection strategies and rely on computationally expensive second-order approximations. In contrast, we demonstrate that mitigating signal collapse, rather than optimizing weight selection, is key to improving accuracy of pruned networks. We propose \textbf{REFLOW} that addresses signal collapse without updating trainable weights, revealing high-quality sparse sub-networks within the original parameter space. REFLOW enables magnitude pruning to achieve state-of-the-art performance, restoring ResNeXt-101 accuracy from under 4.1\% to 78.9\% on ImageNet with only 20\% of the weights retained, surpassing state-of-the-art approaches.

\end{abstract}


\section{Introduction}
\label{sec:introduction}
Neural networks are widely used across applications like natural language processing~\cite{Transformers} and computer vision~\cite{alexnet, RESNET}. However, their increasing size, often reaching billions of parameters~\cite{Young2017RecentTI,ecoflap}, presents significant computational and memory challenges, making deployment in resource-constrained environments impractical~\cite{ZeRO}. Network pruning has emerged as a key technique to reduce model complexity and enable faster inference~\cite{SparseDNN, PruneFL, snip, GraSP, SynFlow}.

The increase in neural network sizes~\cite{Young2017RecentTI} and the widespread availability of pre-trained models have shifted the focus of pruning strategies. Gradual pruning, which iteratively removes parameters with retraining~\cite{late_training_gupta, Kusupati, gale2019state, RIGL}, is computationally expensive for large models, often requiring days or weeks of fine-tuning~\cite{CHITA}. This becomes infeasible for modern networks, particularly in settings with limited training data, such as privacy-sensitive or low-data scenarios~\cite{WoodFisher}. As a result, one-shot pruning, which compresses pre-trained models in a single step~\cite{WoodFisher, CBS, CHITA, wanda, SparseGPT}, has emerged as a scalable and efficient alternative, directly addressing the runtime and computational cost of gradual pruning for large-scale models.

Pruning methods fall into two categories: \textbf{magnitude pruning (MP)} and \textbf{impact-based pruning (IP)}. MP~\cite{Comparing_Biases, Using_Relevance, hanprune, gordon2020compressing} removes small-magnitude weights, but may not be effective as magnitude alone may not reliably capture parameter importance~\cite{CHITA}. IP methods, such as Optimal Brain Damage~\cite{brain_damage}, Optimal Brain Surgeon~\cite{brain_surgeon}, and their modern extensions, use loss-aware weight selection via second-order approximations to identify and remove low-impact weights, followed by weight updates to compensate for loss. Despite outperforming MP, these methods are computationally expensive as they rely on second-order approximations to select which weights to prune.

This raises an important question:
\textit{What drives the higher pruning quality produced by IP methods compared to MP?}

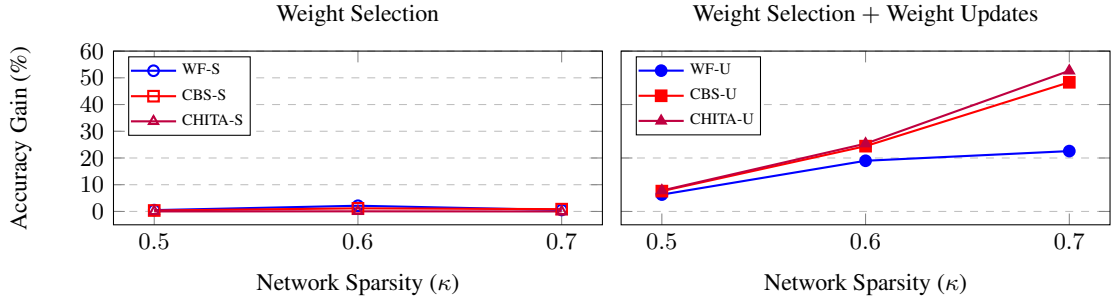
\begin{figure}[t]
\centering
\begin{tikzpicture}

\begin{axis}[
    name=plot1,
    width=6.5cm,
    height=0.14\textwidth,
    scale only axis,
    xlabel={Network Sparsity (\(\kappa\))},
    ylabel={Accuracy Gain (\%)},
    xmin=0.48, xmax=0.72,
    ymin=-5, ymax=60,
    xtick={0.5,0.6,0.7},
    ytick={0,10,20,30,40,50,60},
    ymajorgrids=true,
    grid style=dashed,
    legend pos=north west,
    legend style={font=\tiny, cells={anchor=west}, inner sep=0.5pt},
    tick label style={font=\footnotesize},
    label style={font=\footnotesize},
    title={\footnotesize Weight Selection}
]

\addplot[color=blue, mark=o, solid, thick] coordinates {
    (0.5, 0.49)
    (0.6, 2.13)
    (0.7, 0.55)
};
\addlegendentry{WF-S}

\addplot[color=red, mark=square, solid, thick] coordinates {
    (0.5, 0.35)
    (0.6, 1.16)
    (0.7, 0.85)
};
\addlegendentry{CBS-S}

\addplot[color=purple, mark=triangle, solid, thick] coordinates {
    (0.5, 0.01)
    (0.6, 0.04)
    (0.7, 0.00)
};
\addlegendentry{CHITA-S}

\end{axis}

\begin{axis}[
    name=plot2,
    at={(plot1.east)},
    anchor=west,
    xshift=0.25cm,
    width=6.5cm,
    height=0.14\textwidth,
    scale only axis,
    xlabel={Network Sparsity (\(\kappa\))},
    xmin=0.48, xmax=0.72,
    ymin=-5, ymax=60,
    xtick={0.5,0.6,0.7},
    yticklabels={,,},
    ymajorgrids=true,
    grid style=dashed,
    legend pos=north west,
    legend style={font=\tiny, cells={anchor=west}, inner sep=0.5pt},
    tick label style={font=\footnotesize},
    label style={font=\footnotesize},
    title={\footnotesize Weight Selection $+$ Weight Updates}
]

\addplot[color=blue, mark=*, solid, thick] coordinates {
    (0.5, 6.3)
    (0.6, 18.96)
    (0.7, 22.58)
};
\addlegendentry{WF-U}

\addplot[color=red, mark=square*, solid, thick] coordinates {
    (0.5, 7.6)
    (0.6, 24.43)
    (0.7, 48.33)
};
\addlegendentry{CBS-U}

\addplot[color=purple, mark=triangle*, solid, thick] coordinates {
    (0.5, 7.81)
    (0.6, 25.36)
    (0.7, 52.62)
};
\addlegendentry{CHITA-U}

\end{axis}

\end{tikzpicture}
\caption{Comparison of test accuracy gain of impact-based pruning methods over magnitude pruning for a pre-trained MobileNet on ImageNet at different sparsity levels. \textbf{Left:} Selection-only pruning methods. \textbf{Right:} Pruning methods with weight updates show significant performance gains.}
\label{fig:gain_over_mp_split}
\end{figure}

It is natural to attribute the success of IP methods to \textbf{loss-aware weight selection}, which identifies and preserves critical parameters for the network's performance \cite{brain_damage}. 
\textbf{To test this assumption}, we systematically decouple the two stages of IP: (i) \textbf{weight selection} to determine which weights to prune, and (ii) \textbf{Hessian-based weight update(s)} to adjust the remaining weights and reduce the impact on loss after pruning. Surprisingly, our analysis highlights that \textbf{weight selection alone contributes minimally} to the final performance. Variants of IP-based methods, namely WoodFisher, CBS, and CHITA, that only select weights (WoodFisher-S, CBS-S, CHITA-S) achieve comparable accuracy to MP (Figure~\ref{fig:gain_over_mp_split}, left). The substantial performance improvement is achieved only after Hessian-based updates (Figure~\ref{fig:gain_over_mp_split}, right).

These findings challenge the conventional assumption and leads to a further question:
\textit{If weight selection—whether based on magnitude or loss-aware criteria—plays a minimal role, what fundamentally drives the performance degradation of one-shot pruned networks?}

Our investigation uncovers a new explanation: \textbf{signal collapse}. One-shot pruning alters the flow of activations through the network, progressively reducing their variance in deeper layers. This leads to signal collapse, where activations in the final layers become nearly constant, preventing the network from distinguishing between inputs. As a result, predictions become uniform, severely impairing performance. While accuracy loss due to pruning is conventionally attributed to the removal of critical parameters, our results reveal that signal collapse is the primary driver of degraded performance in pruned networks.

The key insight from our work is that \textbf{signal collapse can be mitigated} and simple methods, such as MP, can surpass the performance of state-of-the-art IP-based methods. Our work \textbf{REFLOW} (\textbf{Re}storing \textbf{F}low of \textbf{Low}-variance signals) mitigates signal collapse without requiring gradient or Hessian-based computations. REFLOW addresses the overlooked bottleneck of signal collapse, demonstrating that high-performing sparse sub-networks inherently exist within the original parameter space identified.

REFLOW achieves substantial accuracy recovery when applied to networks with various neural network architectures pruned with MP. For instance, at 80\% sparsity on ImageNet, ResNet-152 recovers from under 1\% to 68.2\%, and ResNeXt-101 improves from less than 4.1\% to 78.9\%. These results highlight that \textbf{high-quality sparse models} can be obtained by restoring activation flow rather than optimizing weight selection. REFLOW further reveals that high-performing sparse networks inherently exist within the original weights.

\noindent\textbf{Contributions.} This work makes the following key observations and contributions: \begin{enumerate} 
\item 
For the first time, we identify \textbf{signal collapse as the leading cause of accuracy loss in addition to the removal of critical weights,}. 
\item 
\textbf{Signal collapse can be mitigated without updating any trainable weights.} Our work
REFLOW restores activation flow, enabling networks pruned by MP to outperform IP methods \textit{without gradient or Hessian computations}.
\item 
We demonstrate that \textbf{high-performing sparse sub-networks inherently exist in the original parameter space}. Unlike IP methods, which rely on updating unpruned weights to find a solution outside the original parameter space, our approach addresses signal collapse to uncover these sub-networks directly within the original weights. \end{enumerate}

\section{Background \& Related Work}
\label{sec:relatedwork}
This section provides the mathematical formulation of pruning and reviews existing work on pruning techniques.

\subsection{Problem Setup}

Consider a pre-trained deep neural network (DNN) $f(\theta;x)$ parameterized by $\theta \in \mathbb{R}^d$ and input $x$. Pruning produces a sparse sub-network $f(\theta \odot m; x)$, where $m \in \{0,1\}^d$ is a binary mask, and $\odot$ denotes element-wise multiplication. Sparsity $\kappa \in [0,1]$ is the proportion of parameters set to zero.
Pruning assigns scores \( z \in \mathbb{R}^d \) to parameters importance, using methods ranging from simple weight magnitude to loss-aware based pruning scores.

\subsection{Related Work}

\textbf{Magnitude-Based Pruning (MP)}  
is a simple and widely used pruning strategy~\cite{hanprune, lth, Using_Relevance, li2017pruning, SynFlow, Renda2020Comparing, gordon2020compressing, Comparing_Biases, liu2021sparse, DNNShifter}. MP ranks weights based on their absolute values:
\begin{equation}
z_i = |\overline{\theta}_i|.
\label{eq:mp_score}
\end{equation}
It prunes parameters with the smallest magnitudes, which is computationally efficient. However, MP does not account for the impact of pruning on the loss function, which can result in suboptimal pruning decisions.

\textbf{Impact-Based Pruning (IP)}  explicitly considers the loss function to guide pruning decisions \cite{brain_damage, OBS, WoodFisher}. The impact of pruning is quantified as a second-order Taylor expansion of the loss function $\mathcal{L}$ centered at the pre-trained weights $\overline{\theta}$:
\begin{equation}
\mathcal{L}(\overline{\theta} + \delta\theta)-\mathcal{L}(\overline{\theta}) 
= \delta\theta^\top \nabla \mathcal{L}(\overline{\theta})
+ \tfrac{1}{2}\delta\theta^\top H \delta\theta + O(\|\delta\theta\|^3),
\label{eq:taylor_expansion}
\end{equation}
where $H = \nabla^2 \mathcal{L}(\overline{\theta})$ is the Hessian.

Assuming $\overline{\theta}$ represents a local minimum of the loss (as is often the case for pre-trained networks), the gradient term $\nabla \mathcal{L}(\overline{\theta}) = 0$. For small perturbations $\delta\theta$, the higher-order terms become negligible, leading to the local quadratic approximation:
\begin{equation}
\mathcal{L}(\overline{\theta} + \delta\theta)-\mathcal{L}(\overline{\theta}) \approx \tfrac{1}{2}\delta\theta^\top H \delta\theta.
\label{eq:quadratic_approximation}
\end{equation}

Below we review key IP methods that build on this quadratic approximation.

\textit{Optimal Brain Damage (OBD)} improves on MP by explicitly estimating the increase in loss due to pruning~\cite{brain_damage}. Assuming the Hessian $H$ is diagonal, the pruning score for a weight $\overline{\theta}_i$ is:
\begin{equation}
z_i = \frac{\overline{\theta}_i^2}{2H_{ii}}.
\label{eq:obd_score}
\end{equation}
OBD ranks weights based on their impact on loss, using a diagonal Hessian approximation, but ignores parameter interactions.

\textit{Optimal Brain Surgeon (OBS)} generalizes OBD by considering the full Hessian to capture cross-parameter interactions~\cite{brain_surgeon}: 
\begin{equation}
z_i = \frac{\overline{\theta}_i^2}{2[H^{-1}]_{ii}}, \quad 
\delta\theta^{*} = \frac{-\overline{\theta}_i [H^{-1}] e_i}{[H^{-1}]_{ii}}.
\label{eq:obs_score_update}
\end{equation}
Here, $z_i$ represents the pruning score, and $\delta\theta^{*}$ defines the Hessian-based weight updated applied to the unpruned weights. OBS is computationally expensive for modern networks due to the cost of inverting the Hessian $H$; nonetheless, it outperforms MP and OBD.

\textit{Modern Hessian-Based Methods:}  
To reduce the computational cost of OBS, WoodFisher~\cite{WoodFisher} introduces block-diagonal approximations of the Hessian via the empirical Fisher information matrix derived from a subset of training data:
\begin{equation}
H \approx \frac{1}{n}\sum_{i=1}^n \nabla \ell_i(\overline{\theta})\nabla \ell_i(\overline{\theta})^\top,
\label{eq:fisher_approximation}
\end{equation}
where $\ell_i(\overline{\theta})$ is the loss for a single data point. This approximation reduces computational overhead but still focuses on pruning individual weights, without explicitly accounting for interactions between multiple weights.

\textit{Pruning Multiple Weights:} 
Combinatorial Brain Surgeon (CBS)~\cite{CBS} considers the joint effect of pruning multiple weights simultaneously, outperforming WoodFisher. However, its reliance on a dense Hessian $H \in \mathbb{R}^{p \times p}$ makes it computationally intensive, taking hours to prune MobileNet and is not scalable for large networks, such as ResNet-50.
CHITA~\cite{CHITA} uses memory-efficient quadratic approximations for faster pruning than CBS but still relies on Hessian-based updates, modifying unpruned weights rather than identifying existing sparse sub-networks in the original parameter space.

\section{Reassessing Impact-based Pruning}
\label{sec:revisiting_weight_selection}
\subsection{Revisiting Weight Selection}
As discussed in Section~\ref{sec:relatedwork}, MP selects weights based on their absolute magnitudes, while IP's weight selection leverages second-order approximations of the loss (see Equation~\ref{eq:obs_score_update}),  followed by Hessian-based weight updates. To evaluate the role of weight selection in pruning, we compare MP with variants of IP methods, such as WF-S, CBS-S, and CHITA-S (referred to as \textit{IP-selection}), which only prune weights (no weight updates). For additional context, we include random pruning and MP as naive baselines.

Figure~\ref{fig:obs_selection} shows that IP-selection (WF-S, CBS-S, CHITA-S) offers only marginal improvements (up to 2\%) over MP, while random pruning severely reduces accuracy. This indicates that both MP and IP-selection identify meaningful parameters, unlike random pruning. However, the negligible difference between MP and IP-selection underscores the limited role of weight selection in pruning performance.

\begin{figure}[ht]
\centering
\begin{tikzpicture}
\begin{axis}[
    width=10.5cm,
    height=0.15\textwidth,
    scale only axis,
    xlabel={Network Sparsity (\(\kappa\))},
    ylabel={Accuracy \\ Gain (\%)},
    ylabel style={align=center},
    ymin=-65, ymax=5,
    ymajorgrids=true,
    grid style=dashed,
    legend pos=south east,
    legend style={
        at={(0.5,+1.3)},
        anchor=north,
        font=\tiny,
        cells={anchor=west},
        inner sep=2pt,
        legend columns=4,
    },
    tick label style={font=\footnotesize},
    label style={font=\footnotesize},
    legend cell align=left,
    mark options={scale=1},
    cycle list name=color list
]

\addplot[color=blue, mark=o, solid, thick] coordinates {
    (0.5, 0.49)
    (0.6, 2.13)
    (0.7, 0.55)
};
\addlegendentry{WF-S}

\addplot[color=red, mark=square, solid, thick] coordinates {
    (0.5, 0.35)
    (0.6, 1.16)
    (0.7, 0.85)
};
\addlegendentry{CBS-S}

\addplot[color=purple, mark=triangle, solid, thick] coordinates {
    (0.5, 0.01)
    (0.6, 0.04)
    (0.7, 0.00)
};
\addlegendentry{CHITA-S}

\addplot[color=orange, mark=square*, solid, thick] coordinates {
    (0.5, -62.5)
    (0.6, -41.84)
    (0.7, -6.68)
};
\addlegendentry{Random}

\end{axis}
\end{tikzpicture}
\caption{Comparison of test accuracy gain over magnitude pruning for a pre-trained MobileNet (trained on ImageNet) at different sparsity levels.}
\label{fig:obs_selection}
\end{figure}

Further analysis of the similarity between pruning decisions made by MP and CHITA is provided in Appendix~\ref{appendix:pruning_similarity} to demonstrate that both methods produce nearly identical masks, underscoring the limited role of weight selection.

\subsection{Role of Hessian-Based Weight Updates}

While weight selection has negligible impact on accuracy, Hessian-based updates are critical for recovering accuracy by adjusting the remaining weights to compensate for accuracy loss due to pruning.

IP methods combine weight selection with Hessian-based updates (WF-U, CBS-U, CHITA-U). To evaluate the role of updates, we apply Hessian-based updates to MP and the resultant is denoted as MP-U. MP-U tests whether the benefits of Hessian-based updates generalize to MP's simpler selection strategy.
As shown in Figure~\ref{fig:obs_update}, MP-U achieves accuracy gains comparable to WF-U, CBS-U, and CHITA-U. This demonstrates that Hessian-based updates, not weight selection, is the primary driver of accuracy recovery. Combining Hessian-based updates with MP achieves performance on par with state-of-the-art pruning methods, eliminating the need for computationally expensive IP-selection strategies.

\begin{figure}[ht]
\centering
\begin{tikzpicture}
\begin{axis}[
    width=10.5cm,
    height=0.15\textwidth,
    scale only axis,
    xlabel={Network Sparsity (\(\kappa\))},
    ylabel={Accuracy \\ Gain (\%)},
    ylabel style={align=center},
    ymin=-65, ymax=65,
    ymajorgrids=true,
    grid style=dashed,
    legend pos=south east,
    legend style={
        at={(0.5,+1.3)},
        anchor=north,
        font=\tiny,
        cells={anchor=west},
        inner sep=2pt,
        legend columns=5,
    },
    tick label style={font=\footnotesize},
    label style={font=\footnotesize},
    legend cell align=left,
    mark options={scale=1},
    cycle list name=color list
]

\addplot[color=orange, mark=square*, solid, thick] coordinates {
    (0.5, -62.5)
    (0.6, -41.84)
    (0.7, -6.68)
};
\addlegendentry{Random}

\addplot[color=blue, mark=o, solid, thick] coordinates {
    (0.5, 6.3)
    (0.6, 18.96)
    (0.7, 22.58)
};
\addlegendentry{WF-U}

\addplot[color=red, mark=square, solid, thick] coordinates {
    (0.5, 7.6)
    (0.6, 24.43)
    (0.7, 48.33)
};
\addlegendentry{CBS-U}

\addplot[color=teal, mark=halfcircle, solid, thick] coordinates {
    (0.5, 7.63)
    (0.6, 24.39)
    (0.7, 48.73)
};
\addlegendentry{MP-U}

\addplot[color=purple, mark=triangle, solid, thick] coordinates {
    (0.5, 7.63)
    (0.6, 24.39)
    (0.7, 53.4)
};
\addlegendentry{CHITA-U}

\end{axis}
\end{tikzpicture}
\caption{Comparison of test accuracy gain over magnitude pruning for a pre-trained MobileNet (trained on ImageNet) at different sparsity levels.}
\label{fig:obs_update}
\end{figure}

\textbf{Insights}.
IP-selection only methods (WF-S, CBS-S, CHITA-S) offer minimal improvements over MP, confirming that weight selection has little influence on pruning performance. In contrast, Hessian-based update is the primary contributor to accuracy recovery post-pruning. These findings \textit{shift the focus from weight selection to identifying other factors affecting pruning performance}, which is explored in the next section.

\section{Understanding Signal Collapse and Restoring Performance Loss with REFLOW}
\label{sec:Signal_Propagation}
We examine the performance loss of pruned networks by introducing signal collapse - a phenomenon we observe for the first time, where one-shot pruning progressively reduces activation variance across layers, ultimately impairing the network’s ability to distinguish between inputs. In this section, we formally define signal collapse, explain its mechanisms, and demonstrate its impact on network performance. Finally, we introduce REFLOW, a method to mitigate signal collapse and restore the performance of one-shot pruned networks.
\subsection{Notation and Setup}

Consider a pre-trained neural network \( f(\theta) \), parameterized by \(\theta \in \mathbb{R}^d\). For a given layer \(\ell \in \{1, \dots, L\}\), let the input to layer \(\ell\) be denoted as \(\mathbf{H}_{\ell-1}\). The pre-BatchNorm (pre-BN) activation at layer \(\ell\) is defined as:
\begin{equation}
    \mathbf{X}_\ell = f(\mathbf{H}_{\ell-1}; \theta_\ell),
    \label{eq:pre_bn_signal_formal}
\end{equation}
where \(\theta_\ell\) represents the parameters of layer \(\ell\).

Batch Normalization (BN) normalizes the pre-BN activation \(\mathbf{X}_\ell\) across the batch as follows:
\begin{equation}
    \mathbf{Z}_\ell(n) = 
    \frac{\mathbf{X}_\ell(n) - \mu_\ell}
    {\sqrt{\mathrm{Var}_\ell^{\text{(Orig)}}(\mathbf{X}_\ell) + \epsilon}} \cdot \gamma_\ell + \beta_\ell,
    \label{eq:bn_transform}
\end{equation}
where \(n\) is the index of the batch dimension, \(\mu_\ell\) and \(\mathrm{Var}_\ell^{\text{(Orig)}}(\mathbf{X}_\ell)\) are the running mean and variance of the BN layer, \(\gamma_\ell\) and \(\beta_\ell\) are fixed affine parameters, and \(\epsilon > 0\) is a small constant for numerical stability.

\noindent \textbf{Defining Signal Collapse.}
\emph{Signal collapse} occurs in a pruned network if the variance of the activations reduces significantly in deeper layers compared to the original, unpruned network. Formally, let \(\mathrm{Var}_\ell^{(\text{Pruned})}\) and \(\mathrm{Var}_\ell^{(\text{Orig})}\) denote the variances of activations at layer \(\ell\) in the pruned and original networks, respectively. Signal collapse occurs if:
\begin{equation}
    \lim_{\ell \to L} 
    \frac{\mathrm{Var}_\ell^{(\text{Pruned})}}{\mathrm{Var}_\ell^{(\text{Orig})}}
    \;\to\; 0,
    \label{eq:signal_collapse_definition}
\end{equation}
where \(L\) is the total number of layers.

When the variance ratio approaches zero in deeper layers, the activations become nearly constant, resulting in a loss of distinction between inputs, causing uniform predictions.

\subsection{Why Pruning Causes Signal Collapse}
\label{subsec:mechanisms_signal_collapse}

Signal collapse arises due to two reasons explored below:

\paragraph{A) Activation variance reduces due to weight pruning}.

Pruning zeroes out weights based on a selection criterion, typically removing those with lower scores under the assumption that they contribute less to the network's performance. Formally, for layer \(\ell\), pruning modifies the weights \(\theta_\ell\) to \(\theta_\ell'\), where weights are set to zero if their score \(z_{i}\) falls below a threshold \(\tau\):
\begin{equation}
    W_{\ell,i}' = 
    \begin{cases}
        W_{\ell,i}, & \text{if } z_{i} > \tau \\
        0, & \text{otherwise}
    \end{cases},
    \label{eq:weight_pruning_general}
\end{equation}
where \(z_{i}\) represents the pruning score (e.g., magnitude, impact (loss) based heuristic, as discussed in Section \ref{sec:relatedwork}) for weight \(W_{\ell,i}\), and \(\tau\) is the pruning threshold determined by the desired sparsity level \(\kappa\).

To calculate the variance of the pre-BN activation after pruning, consider the activation in its pruned state:\begin{equation}
\mathbf{X}_\ell' = \sum_{i \in \mathcal{S}} W_{\ell,i}' H_{\ell-1,i},
\end{equation}
where \(\mathcal{S}\) is the set of non-pruned weights. Assuming that the activations \( H_{\ell-1,i} \) are independent and have a zero mean, the covariance terms between different \( H_{\ell-1,i} \) and \( H_{\ell-1,j} \) (for \( i \neq j \)) vanish. Therefore, the variance of \(\mathbf{X}_\ell'\) simplifies to:
\begin{equation}
    \mathrm{Var}_\ell^{\text{(Pruned)}}(\mathbf{X}_\ell') = \sum_{i \in \mathcal{S}} W_{\ell,i}'^2 \cdot \mathrm{Var}(H_{\ell-1,i}),
    \label{eq:pruned_variance}
\end{equation}

This leverages the property that the variance of a sum of independent random variables is the sum of their variances, and the scaling property \(\mathrm{Var}(aX) = a^2 \mathrm{Var}(X)\) for a constant \(a\).

Given that at high sparsity many weights are removed, especially those with lower scores, the sum in Equation~\ref{eq:pruned_variance} diminishes, leading to:
\begin{equation}
    \mathrm{Var}_\ell^{\text{(Pruned)}}(\mathbf{X}_\ell') \ll \mathrm{Var}_\ell^{\text{(Orig)}}(\mathbf{X}_\ell).
    \label{eq:variance_reduction}
\end{equation}

\paragraph{B) Over-Normalization due to BN mismatch}

As shown above, after one-shot pruning, pre-BN activations \(\mathbf{X}_\ell'\) have a lower variance, denoted as \(\mathrm{Var}_\ell^{\text{(Pruned)}}(\mathbf{X}_\ell')\). However, BN layers retain the original running statistics \((\mu_\ell, \mathrm{Var}_\ell^{\text{(Orig)}}(\mathbf{X}_\ell))\) computed prior to pruning. Consequently, the BN transformation for pruned activations is:
\begin{equation}
    \mathbf{Z}_\ell'(n) = 
    \frac{\mathbf{X}_\ell'(n) - \mu_\ell}
    {\sqrt{\mathrm{Var}_\ell^{\text{(Orig)}}(\mathbf{X}_\ell) + \epsilon}} \cdot \gamma_\ell + \beta_\ell.
    \label{eq:pruned_bn_variance}
\end{equation}
Here, both subtracting the mean \(\mu_\ell\) and adding  \(\beta_\ell\) do not affect the variance of $\mathbf{Z}_\ell'$, only the scaling factor \(\frac{\gamma_\ell}{\sqrt{\mathrm{Var}_\ell^{\text{(Orig)}}(\mathbf{X}_\ell) + \epsilon}}\) influences the variance of \(\mathbf{Z}_\ell'\).

Applying the property that \(\mathrm{Var}(aX) = a^2 \mathrm{Var}(X)\) for a constant \(a\), the variance of the post-BN activation becomes:
\begin{equation}
    \mathrm{Var}_\ell^{\text{(Pruned)}}(\mathbf{Z}_\ell') = \left( \frac{\gamma_\ell}{\sqrt{\mathrm{Var}_\ell^{\text{(Orig)}}(\mathbf{X}_\ell) + \epsilon}} \right)^2 \cdot \mathrm{Var}_\ell^{\text{(Pruned)}}(\mathbf{X}_\ell').
    \label{eq:var_z_pruned}
\end{equation}
Similarly, the variance of the original (unpruned) post-BN activation is:
\begin{equation}
    \mathrm{Var}_\ell^{\text{(Orig)}}(\mathbf{Z}_\ell) = \left( \frac{\gamma_\ell}{\sqrt{\mathrm{Var}_\ell^{\text{(Orig)}}(\mathbf{X}_\ell) + \epsilon}} \right)^2 \cdot \mathrm{Var}_\ell^{\text{(Orig)}}(\mathbf{X}_\ell).
    \label{eq:var_z_orig}
\end{equation}

Taking the ratio of the pruned to original post-BN variances:
\begin{equation}
    \frac{\mathrm{Var}_\ell^{\text{(Pruned)}}(\mathbf{Z}_\ell')}{\mathrm{Var}_\ell^{\text{(Orig)}}(\mathbf{Z}_\ell)} = \frac{\mathrm{Var}_\ell^{\text{(Pruned)}}(\mathbf{X}_\ell')}{\mathrm{Var}_\ell^{\text{(Orig)}}(\mathbf{X}_\ell)}.
    \label{eq:variance_ratio}
\end{equation}
Given that \(\mathrm{Var}_\ell^{\text{(Pruned)}}(\mathbf{X}_\ell') \ll \mathrm{Var}_\ell^{\text{(Orig)}}(\mathbf{X}_\ell)\) as established in Equation~\ref{eq:variance_reduction}, it follows that:
\begin{equation}
\mathrm{Var}_\ell^{\text{(Pruned)}}(\mathbf{Z}_\ell') \ll \mathrm{Var}_\ell^{\text{(Orig)}}(\mathbf{Z}_\ell).\label{eq:Z_diff}
\end{equation}
Thus, \textbf{Over-Normalization} causes post-BN activations to cluster closely around their mean, reducing their variance. 

\subsection{Cumulative Reduction in Activation Variance and Signal Collapse Across Layers}

Signal collapse arises because of the reduction in activation variance which \emph{cumulatively compounds} across deeper layers in the network. In this subsection, we analyze how this progressive decline emerges when considering post-BN outputs.

\noindent
\textbf{Scaling Factor \(\eta_\ell\).}
For each layer \(\ell\), define the \emph{variance scaling factor} \(\eta_\ell\) as:
\begin{equation}
    \eta_\ell 
    \;=\; 
    \frac{\mathrm{Var}_\ell^{\text{(Pruned)}}\!\bigl(\mathbf{Z}_\ell'\bigr)}
         {\mathrm{Var}_\ell^{\text{(Orig)}}\!\bigl(\mathbf{Z}_\ell\bigr)}
    \;<\; 1,
    \label{eq:scaling_factor}
\end{equation}
where \(\kappa \in [0,1]\) is the fraction of weights pruned. As \(\kappa \to 1\), more weights are removed, which generally makes \(\eta_\ell\) smaller via Equations \ref{eq:pruned_variance} and \ref{eq:Z_diff}.

\noindent
\textbf{Propagation of Variance Reduction.}
Because the \emph{input} to layer \(\ell+1\) is the \emph{post-BN} output of layer \(\ell\) (\(\mathbf{H}_{\ell+1} = \mathbf{Z}_\ell\)), the pruned input \(\mathbf{H}_{\ell+1}'\) is \(\mathbf{Z}_\ell'\). Consequently, any variance reduction at layer \(\ell\) affects layer \(\ell+1\), causing variance to shrink \emph{layer by layer} in the pruned network.

\noindent
\textbf{Cumulative Variance Scaling.}
Recursively applying Equation~\ref{eq:scaling_factor} across \(L\) layers, the cumulative variance scaling at the final layer $L$ becomes:
\begin{equation}
    \mathrm{Var}_L^{\text{(Pruned)}}\!\bigl(\mathbf{Z}_L'\bigr)
    \;=\;
    \Bigl(\,\prod_{\ell=1}^L \eta_\ell\Bigr)
    \,\mathrm{Var}_L^{\text{(Orig)}}\!\bigl(\mathbf{Z}_L\bigr).
    \label{eq:cumulative_variance}
\end{equation}
As an example consider the following. 
By assuming a modest constant per-layer scaling factor of \(\eta_\ell = 0.9\) across \(L = 25\) layers, the cumulative variance scaling becomes:
\begin{equation}
    \prod_{\ell=1}^{25} 0.9 = 0.9^{25} \approx 0.072.
    \label{eq:example_cumulative_variance}
\end{equation}

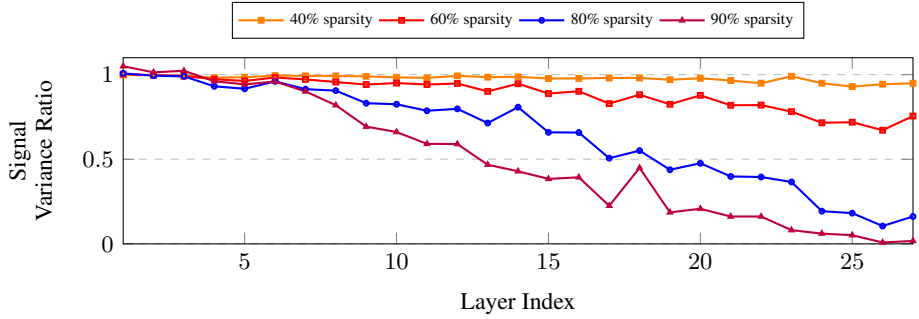
\begin{figure}[h]
\centering
\begin{tikzpicture}
\begin{axis}[
    width=10.5cm,
    height=0.15\textwidth,
    scale only axis,
    xlabel={Layer Index},
    ylabel={Signal \\Variance Ratio},
    ylabel style={align=center},
    xmin=1, xmax=27,
    ymin=0, ymax=1.1,
    xtick={5,10,15,20,25},
    ymajorgrids=true,
    grid style=dashed,
    legend style={
        at={(0.5,+1.3)},
        anchor=north,
        font=\tiny,
        cells={anchor=west},
        inner sep=2pt,
        legend columns=4,
    },
    tick label style={font=\footnotesize},
    label style={font=\footnotesize},
    legend cell align=left,
    cycle list name=color list
]
\addplot[color=orange, mark=square*, mark size=0.9pt, solid, thick] coordinates {
    (1,0.9982) (2,0.9968) (3,0.9934) (4,0.9820) (5,0.9844) (6,0.9947)
    (7,0.9928) (8,0.9922) (9,0.9898) (10,0.9843) (11,0.9805) (12,0.9928)
    (13,0.9851) (14,0.9870) (15,0.9771) (16,0.9775) (17,0.9806) (18,0.9806)
    (19,0.9700) (20,0.9784) (21,0.9651) (22,0.9490) (23,0.9904) (24,0.9490)
    (25,0.9293) (26,0.9433) (27,0.9487)
};

\addplot[color=red, mark=square, mark size=0.9pt, solid, thick] coordinates {
    (1,1.0035) (2,0.9968) (3,0.9921) (4,0.9729) (5,0.9626) (6,0.9830) 
    (7,0.9717) (8,0.9561) (9,0.9413) (10,0.9502) (11,0.9416) (12,0.9476) 
    (13,0.9010) (14,0.9464) (15,0.8883) (16,0.9015) (17,0.8289) (18,0.8813) 
    (19,0.8248) (20,0.8777) (21,0.8190) (22,0.8200) (23,0.7814) (24,0.7156) 
    (25,0.7188) (26,0.6714) (27,0.7553)
};

\addplot[color=blue, mark=o, mark size=0.9pt, solid, thick] coordinates {
    (1,1.0071) (2,0.9936) (3,0.9898) (4,0.9310) (5,0.9164) (6,0.9603) 
    (7,0.9142) (8,0.9054) (9,0.8313) (10,0.8248) (11,0.7865) (12,0.7979) 
    (13,0.7139) (14,0.8078) (15,0.6590) (16,0.6577) (17,0.5062) (18,0.5508) 
    (19,0.4380) (20,0.4762) (21,0.3983) (22,0.3949) (23,0.3657) (24,0.1931) 
    (25,0.1815) (26,0.1053) (27,0.1616)
};

\addplot[color=purple, mark=triangle, mark size=0.9pt, solid, thick] coordinates {
    (1,1.0497) (2,1.0136) (3,1.0227) (4,0.9611) (5,0.9423) (6,0.9593)
    (7,0.9009) (8,0.8193) (9,0.6927) (10,0.6609) (11,0.5910) (12,0.5896)
    (13,0.4677) (14,0.4287) (15,0.3839) (16,0.3930) (17,0.2250) (18,0.4487)
    (19,0.1859) (20,0.2073) (21,0.1615) (22,0.1616) (23,0.0810) (24,0.0605)
    (25,0.0511) (26,0.0081) (27,0.0171)
};
\addlegendentry{40\% sparsity}
\addlegendentry{60\% sparsity}
\addlegendentry{80\% sparsity}
\addlegendentry{90\% sparsity}
\end{axis}
\end{tikzpicture}
\caption{Layer-wise signal variance ratios $
    \frac{\mathrm{Var}^{\text{(Pruned)}}}{\mathrm{Var}^{\text{(Orig)}}},
    \label{eq:variance_ratio}$ in pruned MobileNet (on ImageNet). Higher sparsity leads to severe signal collapse in deeper layers.}\label{fig:variance_ratios}
\end{figure}

In practice, \(\eta_\ell\) varies across layers due to differences in sparsity distribution across layers, but the overall trend remains consistent: variance reduces as activations propagate through the pruned network. As \(\ell\) approaches \(L\), with increasing sparsity  \(\kappa\), 
\begin{equation}
    \;\;\lim_{\kappa \to 1}\;
    \mathrm{Var}_L^{\text{(Pruned)}}\bigl(\mathbf{Z}_L'\bigr)
    \;=\;
    \Bigl(\prod_{\ell=1}^L \eta_\ell\Bigr)\,
    \mathrm{Var}_L^{\text{(Orig)}}\!\bigl(\mathbf{Z}_L\bigr)
    \;\;\xrightarrow{}\;\;
    0,
    \label{eq:variance_approaching_zero}
\end{equation}
Consequently, the variance of the activations in the final layers approaches zero (see Figure~\ref{fig:variance_ratios}). 

\begin{figure}[t]
\centering
\begin{tikzpicture}
\begin{axis}[
    width=10.5cm,
    height=0.15\textwidth,
    scale only axis,
    xlabel={Class Index},
    ylabel={Fraction of Predictions},
    xmin=-0.5, xmax=9.5,
    ymin=0, ymax=1.1,
    xtick={0,2,4,6,8},
    ymajorgrids=true,
    grid style=dashed,
    legend style={
        at={(0.02,0.98)},
        anchor=north west,
        font=\tiny,
        cells={anchor=west},
        inner sep=0.5pt,
        legend columns=1,
        scale=0.65,
        row sep=-2pt,
    },
    tick label style={font=\footnotesize},
    label style={font=\footnotesize},
    title={\footnotesize Distribution of Predictions},
    ybar,
    bar width=3,
]
\addplot[blue!60, fill=blue!60] coordinates {
    (0,0.099600) (1,0.101300) (2,0.100100) (3,0.101500) (4,0.099300)
    (5,0.098900) (6,0.100600) (7,0.099900) (8,0.098700) (9,0.100100)
};
\addlegendentry{Unpruned}

\addplot[red!60, fill=red!60] coordinates {
    (0,0.000000) (1,0.000000) (2,0.000500) (3,0.000000) (4,0.000000)
    (5,0.000000) (6,0.000000) (7,0.994800) (8,0.000000) (9,0.004700)
};
\addlegendentry{90\% Sparsity}
\end{axis}
\end{tikzpicture}
\caption{Distribution of predictions made by ResNet-20 on CIFAR-10. The unpruned model predicts uniformly across classes, discriminating between inputs, while the pruned model maps most inputs to a single class.}
\label{fig:prediction_collapse}
\end{figure}
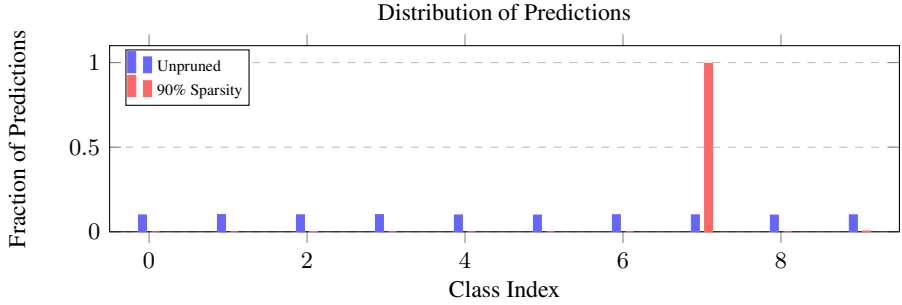

\noindent
\textbf{Insight.}
Given that \(\mathrm{Var}_L^{\text{(Pruned)}}(\mathbf{Z}_L') \to 0\), the post-BN outputs converge to their mean (constant) value:
\begin{equation}
    \lim_{\mathrm{Var}(\mathbf{Z}_L') \to 0} 
    \;\mathbf{Z}_L'(n) 
    \;=\; 
    \mathrm{Mean}(\mathbf{Z}_L'(n))
    \label{eq:constant_activation}
\end{equation}
This implies that for any two distinct inputs \(\mathbf{x}_1\) and \(\mathbf{x}_2\),
$\mathbf{Z}_L'(\mathbf{x}_1) \approx \mathbf{Z}_L'(\mathbf{x}_2)$. We refer to this as \textit{signal collapse} (see Figure~\ref{fig:prediction_collapse}) and the network is incapable of distinguishing between inputs, thereby making it ineffective for downstream tasks, such as classification.

We empirically validate signal collapse using two \textbf{global scalar metrics} after each BN layer \(\ell\). These metrics summarize the behavior of normalized activations \(\mathbf{Z}_\ell\),

\begin{align}
    \mathrm{Mean}_\ell &= \frac{1}{|\mathbf{Z}_\ell|} \sum_{x \in \mathbf{Z}_\ell} x, \\
    \mathrm{Var}_\ell &= \frac{1}{|\mathbf{Z}_\ell|} \sum_{x \in \mathbf{Z}_\ell} \left( x - \mathrm{Mean}_\ell \right)^2.
\end{align}
where \(|\mathbf{Z}_\ell|\) denotes the total number of elements in the tensor \(\mathbf{Z}_\ell\). Figure~\ref{fig:variance_ratios} illustrates the signal variance ratio \(\frac{\mathrm{Var}_\ell^{\text{(Pruned)}}}{\mathrm{Var}_\ell^{\text{(Orig)}}}\) across layers at different sparsity levels \(\kappa\). The ratio progressively drops with increasing sparsity, remaining near 1 for 40\% sparsity, but drops below 0.1 after Layer~22 in 90\% sparsity, resulting in signal collapse.

Figure~\ref{fig:prediction_collapse} demonstrates signal collapse in ResNet-20 on CIFAR-10. The unpruned model distributes predictions evenly across all classes in the test set. In contrast, at 90\% sparsity, the pruned network collapses, with over 99.4\% of inputs mapped to a single class. This behavior aligns with our analysis shown in Equation~\ref{eq:constant_activation}, leading to nearly identical representations for different inputs ($\mathbf{Z}'_L(\mathbf{x}_1) \approx \mathbf{Z}'_L(\mathbf{x}_2)$).

\subsection{Hessian-Based Updates Mitigate but Do Not Fully Solve Signal Collapse}

Building on our earlier findings that Hessian-based updates are essential to recovering accuracy after pruning, we examine the effect of these updates on signal collapse. Specifically, we analyze how these updates influence the variance of activations across network layers.

\paragraph{Impact of Hessian-Based Updates on Signal Variance}

\begin{figure}[h]
\centering
\begin{tikzpicture}
\begin{axis}[
    width=10.5cm,
    height=0.15\textwidth,
    scale only axis,
    xlabel={Layer Index},
    ylabel={Signal \\Variance Ratio},
    ylabel style={align=center},
    xmin=1, xmax=27,
    ymin=0, ymax=1.1,
    xtick={5,10,15,20,25},
    ymajorgrids=true,
    grid style=dashed,
    legend pos=south west,
    legend style={
        font=\tiny,
        cells={anchor=west},
        inner sep=2pt,
        legend columns=3,
    },
    tick label style={font=\footnotesize},
    label style={font=\footnotesize},
    legend cell align=left,
    cycle list name=color list
]
\addplot[color=blue, mark=square*, mark size=0.9pt, solid, thick] coordinates {
    (1,0.995) (2,1.011) (3,1.006) (4,0.956) (5,0.928) (6,0.978)
    (7,0.927) (8,0.925) (9,0.844) (10,0.855) (11,0.804) (12,0.837)
    (13,0.726) (14,0.849) (15,0.678) (16,0.721) (17,0.523) (18,0.634)
    (19,0.448) (20,0.579) (21,0.408) (22,0.523) (23,0.371) (24,0.298)
    (25,0.186) (26,0.208) (27,0.173)
};
\addplot[color=red, mark=square, mark size=0.9pt, solid, thick] coordinates {
    (1,0.982) (2,1.009) (3,1.002) (4,0.953) (5,0.923) (6,0.973)
    (7,0.925) (8,0.926) (9,0.844) (10,0.855) (11,0.807) (12,0.841)
    (13,0.728) (14,0.851) (15,0.680) (16,0.722) (17,0.523) (18,0.635)
    (19,0.448) (20,0.580) (21,0.408) (22,0.523) (23,0.371) (24,0.298)
    (25,0.186) (26,0.208) (27,0.173)
};
\addplot[color=green, mark=o, mark size=0.9pt, solid, thick] coordinates {
    (1,0.998) (2,0.990) (3,0.960) (4,0.979) (5,0.948) (6,0.985)
    (7,0.922) (8,0.942) (9,0.901) (10,0.954) (11,0.827) (12,0.886)
    (13,0.806) (14,0.914) (15,0.616) (16,0.852) (17,0.583) (18,0.804)
    (19,0.534) (20,0.744) (21,0.518) (22,0.694) (23,0.496) (24,0.645)
    (25,0.396) (26,1.008) (27,0.731)
};
\addlegendentry{CHITA-S}
\addlegendentry{MP-S}
\addlegendentry{CHITA-U}
\end{axis}
\end{tikzpicture}
\caption{Layer-wise signal variance ratios $\frac{\mathrm{Var}^{\text{(Pruned)}}}{\mathrm{Var}^{\text{(Baseline)}}}$ in 80\% sparse MobileNet on ImageNet. \textbf{CHITA-S} and \textbf{MP-S} show identical levels of signal collapse, while \textbf{CHITA-U} mitigates this collapse by Hessian-based updates.}\label{fig:variance_updates}
\end{figure}
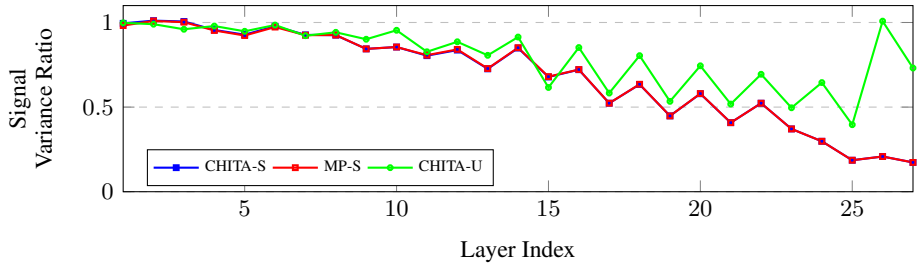

Figure~\ref{fig:variance_updates} presents the layer-wise signal variance ratios \(\frac{\mathrm{Var}_\ell^{\text{(Pruned)}}}{\mathrm{Var}_\ell^{\text{(Orig)}}}\) for three pruning methods: \textbf{CHITA-S} (selection-only), \textbf{MP-S} (Magnitude Pruning selection-only), and \textbf{CHITA-U} (selection with Hessian-based updates). 

As presented in Section~\ref{sec:revisiting_weight_selection}, \textbf{CHITA-S} and \textbf{MP-S} make near similar pruning decisions, resulting in identical (overlapping) variance profiles. Both pruning methods undergo signal collapse, with variance ratios frequently below 1 across layers. In contrast, \textbf{CHITA-U} maintains higher variance ratios in most layers, indicating that Hessian-based updates may help mitigate signal collapse. However, a few layers deeper in the network still show reduced variance ratios. This suggests that while Hessian-based updates reduce the impact of signal collapse, they do not entirely prevent it.

\subsection{REFLOW: Restoring Signal Propagation to Mitigate Collapse}

\emph{Signal collapse} due to pruning stems from diminished activation variance. Hessian-based IP methods only partially mitigate signal collapse by updating the unpruned weights. These observations point to a more direct solution: if the core issue is the compounding mismatch between the pruned and original activation variances \(\frac{\mathrm{Var}_\ell^{\text{(Pruned)}}}{\mathrm{Var}_\ell^{\text{(Orig)}}} < 1\), resulting in signal collapse (Equation~\ref{eq:signal_collapse_definition}), then the running BN statistics can be calibrated to induce activation variance to mitigate signal collapse,  \emph{without} updating any unpruned (trainable) weights.

We propose \textbf{REFLOW} (\textbf{Re}storing \textbf{F}low of \textbf{Low}-variance signals), which updates only the BN running mean and variance after one-shot pruning. Specifically, rather than relying on the pre-pruning BN statistics $\bigl(\mu_\ell,\;\mathrm{Var}_\ell^{(\text{Orig})}\bigr)$, we collect a small calibration set $\mathcal{B}$ ($O(10)$ training batches) and pass it through the pruned network, and compute:
\begin{align}
\widehat{\mu}_\ell^{(\text{Pruned})}
  &= \frac{1}{|\mathcal{B}|} \sum_{n \in \mathcal{B}} \mathbf{X}_\ell'(n), \\
\widehat{\mathrm{Var}}_\ell^{(\text{Pruned})}(\mathbf{X}_\ell')
  &= \frac{1}{|\mathcal{B}|} \sum_{n \in \mathcal{B}}
    \Bigl(\mathbf{X}_\ell'(n) - \widehat{\mu}_\ell^{(\text{Pruned})}\Bigr)^2,
  \label{eq:reflow_calib}
\end{align}
where $\mathbf{X}_\ell'(n)$ are the pre-BN activations in the pruned model. We replace 
$\bigl(\mu_\ell,\;\mathrm{Var}_\ell^{(\text{Orig})}\bigr)$
with $\bigl(\widehat{\mu}_\ell^{(\text{Pruned})},\;\widehat{\mathrm{Var}}_\ell^{(\text{Pruned})}\bigr)$ in the BN layers, leaving $\gamma_\ell,\;\beta_\ell$ and all unpruned weights unchanged. The resulting post-BN activations become:
\begin{equation}
  \mathbf{Z}_\ell'^{(\text{REFLOW})}(n)
  \;=\;
  \frac{\mathbf{X}_\ell'(n) - \widehat{\mu}_\ell^{(\text{Pruned})}}
       {\sqrt{\widehat{\mathrm{Var}}_\ell^{(\text{Pruned})}(\mathbf{X}_\ell') + \epsilon}}
  \;\gamma_\ell \;+\; \beta_\ell.
  \label{eq:reflow_transform}
\end{equation}

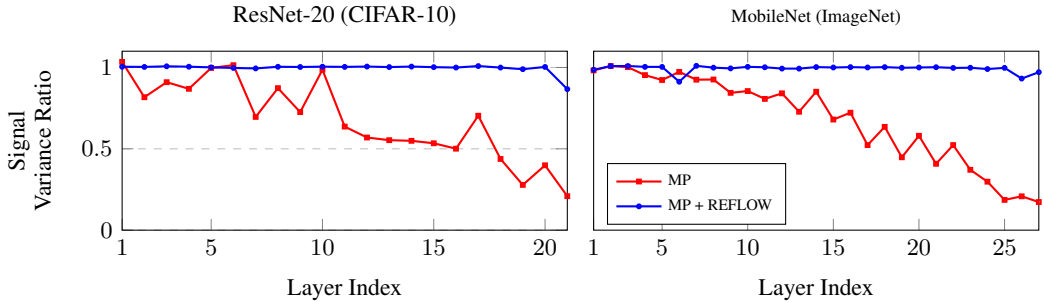
\begin{figure}[h]
\centering
\begin{tikzpicture}
\begin{groupplot}[
    group style={
        group size=2 by 1,
        horizontal sep=10pt,
        vertical sep=0pt
    },
    width=7.5cm,
    height=0.24\textwidth,
    xlabel={Layer Index},
    xmin=1,
    ymin=0, ymax=1.1,
    ymajorgrids=true,
    grid style=dashed,
    tick label style={font=\footnotesize},
    label style={font=\footnotesize},
    legend cell align=left
]

\nextgroupplot[
    xmax=21,
    xtick={1,5,10,15,20},
    ylabel={Signal \\ Variance Ratio},
    ylabel style={align=center},
    title={\footnotesize ResNet-20 (CIFAR-10)}
]
\addplot[color=red, mark=square, mark size=0.7pt, solid, thick] coordinates {
(1,1.0345) (2,0.8172) (3,0.9096) (4,0.8688) (5,0.9972) (6,1.0141) (7,0.6961) (8,0.8730) (9,0.7257) (10,0.9861) (11,0.6367) (12,0.5695) (13,0.5532) (14,0.5492) (15,0.5341) (16,0.5009) (17,0.7034) (18,0.4376) (19,0.2782) (20,0.3987) (21,0.2088)};

\addplot[color=blue, mark=o, mark size=0.7pt, solid, thick] coordinates {
(1,1.0045) (2,1.0033) (3,1.0071) (4,1.0047) (5,1.0007) (6,0.9972) (7,0.9942) (8,1.0043) (9,1.0030) (10,1.0049) (11,1.0036) (12,1.0056) (13,1.0025) (14,1.0055) (15,1.0020) (16,0.9995) (17,1.0086) (18,0.9994) (19,0.9895) (20,1.0033) (21,0.8669)};

\nextgroupplot[
    xmax=27,
    xtick={1,5,10,15,20,25},
    ytick = \empty,
    legend style={
        at={(0.03,0.03)},
        anchor=south west,
        font=\tiny,
        cells={anchor=west},
    },
    title={\scriptsize MobileNet (ImageNet)}
]
\addplot[color=red, mark=square, mark size=0.7pt, solid, thick] coordinates {
    (1,0.982) (2,1.009) (3,1.002) (4,0.953) (5,0.923) (6,0.973)
    (7,0.925) (8,0.926) (9,0.844) (10,0.855) (11,0.807) (12,0.841)
    (13,0.728) (14,0.851) (15,0.680) (16,0.722) (17,0.523) (18,0.635)
    (19,0.448) (20,0.580) (21,0.408) (22,0.523) (23,0.371) (24,0.298)
    (25,0.186) (26,0.208) (27,0.173)
};
\addlegendentry{MP}

\addplot[color=blue, mark=o, mark size=0.7pt, solid, thick] coordinates {
(1,0.9857) (2,1.0085) (3,1.0091) (4,1.0037) (5,1.0035) (6,0.9124) (7,1.0100) (8,0.9989) (9,0.9944) (10,1.0039) (11,1.0013) (12,0.9932) (13,0.9930) (14,1.0024) (15,1.0000) (16,1.0021) (17,1.0000) (18,1.0021) (19,0.9981) (20,1.0000) (21,1.0016) (22,0.9972) (23,0.9983) (24,0.9904) (25,0.9976) (26,0.9318) (27,0.9710)};
\addlegendentry{MP + REFLOW}

\end{groupplot}
\end{tikzpicture}

\caption{Layer-wise signal variance ratios in pruned networks under magnitude pruning at 80\% sparsity, before and after applying REFLOW.}
\label{fig:reflow_effect}
\end{figure}

\smallskip \noindent \textbf{Preserving Signal Variance}: REFLOW aligns each BN layer’s statistics with the \emph{true} statistics of the pruned pre-BN activations. This offsets the cumulative loss of variance that causes signal collapse. Figure~\ref{fig:reflow_effect}, shows recalibration at high sparsities restores the post-BN variance to the levels of the unpruned network. In doing so, REFLOW improves the network’s discriminative power by mitigating signal collapse without retraining or updating unpruned weights.

\section{Experimental Results}
\label{sec:studies}
We apply REFLOW to magnitude pruning (MP) and evaluate it across small, medium, and large architectures. The results highlight REFLOW’s consistently recovers performance in pruned networks, achieving state-of-the-art accuracy without requiring computationally expensive Hessian-based updates. By mitigating signal collapse, REFLOW enables the discovery of high-quality sparse subnetworks within the original parameter space.

\subsection{Performance on Small Architectures}
We begin by evaluating REFLOW on small architectures, namely ResNet-20~\cite{RESNET} pre-trained on CIFAR-10~\cite{CIFAR10} and MobileNet~\cite{MobileNet} pre-trained on ImageNet~\cite{ImageNet}, with less than 5 million parameters and comparing them to state-of-the-art one-shot pruning approaches, namely WF~\cite{WoodFisher}, CBS~\cite{CBS}, and CHITA~\cite{CHITA}. 

Table~\ref{tab:small_architectures_results} highlights REFLOW’s accuracy improvements across all sparsity levels. For ResNet-20, REFLOW restores accuracy to 49.16\% at 0.9 sparsity, outperforming CHITA (15.60\%) and MP (11.79\%). On MobileNet, REFLOW achieves 43.37\% accuracy at 0.8 sparsity, surpassing CHITA (29.78\%) and MP (0.11\%). 

\begin{table}[h]
    \small
    \setlength{\tabcolsep}{4pt} 
    \caption{Performance of pruning methods on small architectures (ResNet-20 on CIFAR-10 and MobileNet on ImageNet) across varying sparsity levels. Sparsity values represent the fraction of weights pruned (e.g., 0.4 corresponds to 40\% pruning). The unpruned test accuracy for ResNet-20 and MobileNet are 91.57\% and 71.96\%, respectively. The best accuracy values are highlighted in bold. Weight update indicates whether a single-pass Hessian-based update is performed on unpruned weights post-pruning.}
    \label{tab:small_architectures_results}
    \centering
    \resizebox{0.7\columnwidth}{!}{%
    \begin{tabular}{cccccccc}
        \toprule
        Dataset & Network & Sparsity & MP & WF & CBS & CHITA & REFLOW \\
        \midrule
        \multirow{6}{*}{CIFAR-10} & \multirow{6}{*}{ResNet-20} 
        & 0.4 & 89.98 & 91.15 & 91.21 & 91.19 & \textbf{91.25} \\
        & & 0.5 & 88.44 & 90.23 & 90.58 & 90.60 & \textbf{90.66} \\
        & & 0.6 & 85.24 & 87.96 & 88.88 & 89.22 & \textbf{89.49} \\
        & & 0.7 & 78.79 & 81.05 & 81.84 & 84.12 & \textbf{86.65} \\
        & & 0.8 & 54.01 & 62.63 & 51.28 & 57.90 & \textbf{78.50} \\
        & & 0.9 & 11.79 & 11.49 & 13.68 & 15.60 & \textbf{49.16} \\
        \midrule
        \multirow{5}{*}{ImageNet} & \multirow{5}{*}{MobileNet} 
        & 0.4 & 69.16 & 71.15 & 71.45 & 71.50 & \textbf{71.59} \\
        & & 0.5 & 62.61 & 68.91 & 70.21 & 70.42 & \textbf{70.48} \\
        & & 0.6 & 41.94 & 60.90 & 66.37 & 67.30 & \textbf{67.83} \\
        & & 0.7 & 6.78 & 29.36 & 55.11 & 59.40 & \textbf{61.54} \\
        & & 0.8 & 0.11 & 0.24 & 16.38 & 29.78 & \textbf{43.37} \\
        \midrule
        Weight Update & - & - &  \xmark & \cmark & \cmark & \cmark & \xmark \\
        \bottomrule
    \end{tabular}
    }
\end{table}

\subsection{Scaling REFLOW to Medium-sized Architectures}
We next evaluate REFLOW on medium-sized architectures, namely ResNet-50 (ImageNet) with less than 25 million parameters. For this size, we compare REFLOW to CHITA and M-FAC~\cite{mfac}, as WF and CBS are computationally prohibitive.

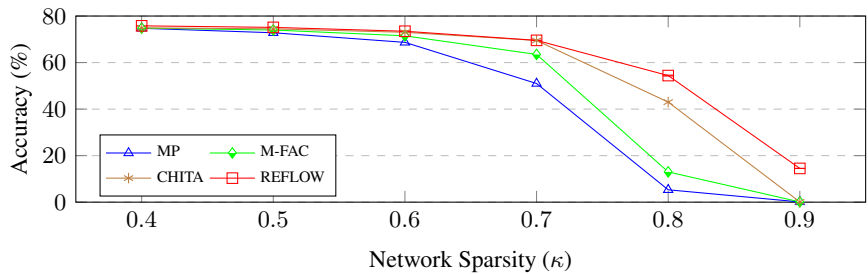
\begin{figure}[h]
    \centering
    \begin{tikzpicture}
    \begin{axis}[
        width=10.5cm,
        height=0.15\textwidth,
        scale only axis,
        xlabel={Network Sparsity (\(\kappa\))},
        ylabel={Accuracy (\%)},
        xmin=0.35, xmax=0.95,
        ymin=0, ymax=80,
        ylabel near ticks,
        ylabel shift=-3pt,
        xtick={0.4,0.5,0.6,0.7,0.8,0.9},
        ymajorgrids=true,
        grid style=dashed,
        legend pos=south west,
        legend style={font=\tiny, cells={anchor=west}, inner sep=2pt, legend columns=2},
        tick label style={font=\footnotesize},
        label style={font=\footnotesize},
        legend cell align=left,
        mark options={scale=1},
        cycle list name=color list
    ]

    \addplot[color=blue,mark=triangle] coordinates {
        (0.40,74.74)
        (0.50,72.81) (0.60,68.68) 
        (0.70,51) (0.80,5.318) 
        (0.90,0.1)
    };
    \addlegendentry{MP}

    \addplot[color=green,mark=halfdiamond*] coordinates {
        (0.40,74.8)
        (0.50,74) (0.60,71.5) 
        (0.70,63.5) (0.80,13) 
        (0.90,0.1)
    };
    \addlegendentry{M-FAC}

    \addplot[color=brown,mark=asterisk] coordinates {
        (0.40,74.9)
        (0.50,74.5) (0.60,73) 
        (0.70,69.5) (0.80,43) 
        (0.90,0.1)
    };
    \addlegendentry{CHITA}

    \addplot[color=red,mark=square, error bars/.cd, y dir=both, y explicit] 
    coordinates {
        (0.40,75.788) +- (0,0.04)
        (0.50,75.12) +- (0,0.03)
        (0.60,73.466) +- (0,0.01)
        (0.70,69.55) +- (0,0.08)
        (0.80,54.412) +- (0,0.19)
        (0.90,14.506) +- (0,0.04)
    };
    \addlegendentry{REFLOW}

    \end{axis}
    \end{tikzpicture}
    \caption{Comparison of test accuracy for varying sparsities across pruning techniques for ResNet-50 on ImageNet. REFLOW outperforms CHITA, M-FAC, and MP consistently.}
    \label{fig:resnet50_results}
\end{figure}

Figure~\ref{fig:resnet50_results} shows that REFLOW outperforms CHITA and M-FAC across all sparsity levels. At high sparsities, REFLOW offer superior accuracy improvements without the overhead of Hessian-based updates. 

\subsection{Scaling REFLOW to Large Architectures}
\begin{figure}[h]
    \centering
    \begin{tikzpicture}
    \begin{axis}[
        width=11.5cm,
        height=0.25\textwidth,
        ybar,
        bar width=8pt,
        xlabel={Architecture},
        ylabel={\footnotesize Accuracy (\%)},
        ylabel style={font=\footnotesize}, 
        ymin=0, ymax=90,
        xtick=data,
        xticklabel style={font=\footnotesize}, 
        symbolic x coords={ResNet-101, ResNet-152, RegNetX, ResNeXt-101},
        grid=both,
        grid style=dashed,
        ymajorgrids=true,
        tick label style={font=\footnotesize}, 
        label style={font=\footnotesize}, 
        ylabel near ticks,
        ylabel shift=-3pt,
        legend style={font=\footnotesize, cells={anchor=west}, inner sep=2pt, legend columns=1, at={(0.5,1.05)}, anchor=south}, 
        nodes near coords,
        every node near coord/.append style={font=\fontsize{0.1}{0.1}\selectfont} 
    ]

    \addplot[fill=blue] coordinates {
        (ResNet-101, 4.1)
        (ResNet-152, 0.9)
        (RegNetX, 1.1)
        (ResNeXt-101, 4.1)
    };

    \addplot[fill=red] coordinates {
        (ResNet-101, 64.1)
        (ResNet-152, 68.2)
        (RegNetX, 73.0)
        (ResNeXt-101, 78.9)
    };

    \end{axis}
    \end{tikzpicture}
    \caption{Accuracy comparison of MP (blue) and REFLOW applied to MP (red) at 80\% sparsity for large architectures pre-trained on ImageNet. REFLOW mitigates signal collapse and restores accuracy.}
    \label{fig:large_architectures}
\end{figure}
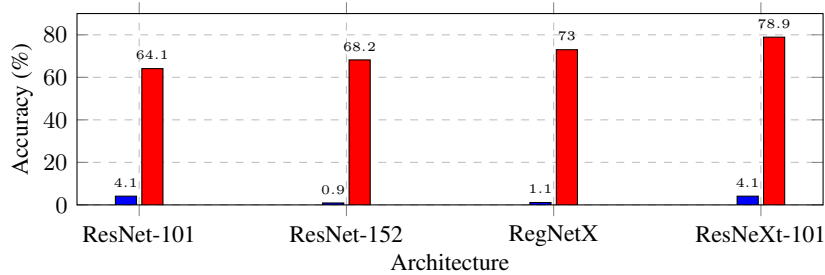

Finally, we consider large architectures, namely ResNet-101, ResNet-152, RegNetX-32GF (with nearly 107 million parameters), and ResNeXt-101 (64x4d) that has over 45 million parameters. These models pose significant challenges for pruning, particularly for CHITA as it relies on memory- and computation-intensive second-order approximations.
Figure~\ref{fig:large_architectures} shows that REFLOW delivers up to 74.8\% accuracy gains over MP on ImageNet. For instance, on ResNet-101, REFLOW restores accuracy from 4.1\% (MP) to 64.1\%. On ResNet-152, REFLOW achieves 68.2\% accuracy, compared to just 0.9\% for MP. Similar gains are observed for RegNetX-32GF, where REFLOW achieves 73.0\% accuracy, and for ResNeXt-101, where it achieves 78.9\%, outperforming MP. 

\subsection{Convergence with REFLOW}
Building on the results in Table~\ref{tab:small_architectures_results}, we evaluate the impact of REFLOW across pruning methods with varying complexities: MP, CHITA-S (selection-only), and CHITA (selection with Hessian-based updates). CHITA updates the unpruned weights using second-order information, while CHITA-S applies the same selection criteria without weight updates. This distinction isolates the role of weight updates and quantifies whether REFLOW can compensate for their absence.

\begin{figure}[h]
    \centering
    \begin{tikzpicture}
    \begin{axis}[
        width=10.5cm,
        height=0.15\textwidth,
        scale only axis,
        xlabel={Network Sparsity (\(\kappa\))},
        ylabel={Accuracy (\%)},
        xmin=0.35, xmax=0.95,
        ymin=0, ymax=100,
        ylabel near ticks,
        ylabel shift=-3pt,
        xtick={0.4,0.5,0.6,0.7,0.8,0.9},
        ymajorgrids=true,
        grid style=dashed,
        legend pos=south west,
        legend style={font=\tiny, cells={anchor=west}, inner sep=2pt, legend columns=2},
        tick label style={font=\footnotesize},
        label style={font=\footnotesize},
        legend cell align=left,
        mark options={scale=1},
        cycle list name=color list
    ]

    \addplot[color=blue,mark=triangle] coordinates {
        (0.3,90.77) (0.4,89.98) (0.5,88.44) 
        (0.6,85.24) (0.7,78.79) (0.8,54.01) 
        (0.9,11.79)
    };
    \addlegendentry{MP}

    \addplot[color=red,mark=diamond*] coordinates {
        (0.3,91.41) (0.4,91.05) (0.5,90.39) 
        (0.6,89.38) (0.7,86.65) (0.8,78.5) 
        (0.9,49.0)
    };
    \addlegendentry{MP + REFLOW}

    \addplot[color=green,mark=asterisk] coordinates {
        (0.3,91.37) (0.4,91.0) (0.5,90.45) 
        (0.6,89.41) (0.7,86.45) (0.8,78.79) 
        (0.9,49.35)
    };
    \addlegendentry{CHITA-S + REFLOW}

    \addplot[color=brown,mark=square, error bars/.cd, y dir=both, y explicit] 
    coordinates {
        (0.3,91.43) (0.4,91.17) (0.5,90.68) 
        (0.6,89.64) (0.7,87.08) (0.8,78.84) 
        (0.9,45.76)
    };
    \addlegendentry{CHITA + REFLOW}

    \end{axis}
    \end{tikzpicture}
    \caption{Comparison of test accuracy vs. sparsity for ResNet-20 on CIFAR-10.}
    \label{fig:reflow_convergence}
\end{figure}
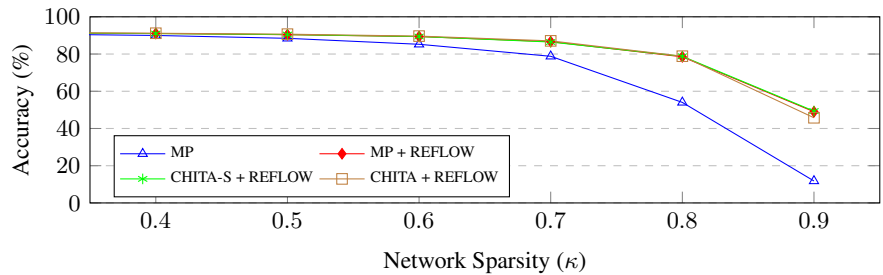

Figure~\ref{fig:reflow_convergence} shows that REFLOW bridges the performance gap between MP, CHITA-S, and CHITA. REFLOW enables simpler selection based approaches like MP and CHITA-S to achieve comparable accuracy as CHITA (Hessian-based weight updates) although the latter is computationally intensive. This highlights that mitigating signal collapse, rather than employing complex pruning selection heuristics, is the key to recovering performance in one-shot pruned networks.

\section{Conclusion}
\label{sec:discussion}
This work identifies signal collapse as a critical bottleneck in one-shot neural network pruning. Performance loss in pruned networks is due to \textbf{signal collapse} in addition to the removal of critical parameters. We propose \textbf{REFLOW} (\textbf{Re}storing \textbf{F}low of \textbf{Low}-variance signals), a simple yet effective method that mitigates signal collapse without computationally expensive weight updates. By focusing on signal preservation, REFLOW highlights the importance of mitigating signal collapse in sparse networks and enables magnitude pruning to match or surpass state-of-the-art one-shot pruning methods such as CHITA, CBS, and WF.

REFLOW consistently achieves state-of-the-art accuracy across diverse architectures, restoring ResNeXt-101 from under 4.1\% to 78.9\% top-1 accuracy at 80\% sparsity on ImageNet. Its lightweight design makes it a practical solution for both research and deployment, delivering high-quality sparse models without the overhead of traditional approaches. These findings challenge the traditional emphasis on weight selection strategies and underscore the critical role of signal propagation for achieving high-quality sparse networks in the context of one-shot pruning.

\bibliographystyle{plainnat}
\bibliography{Main}

\newpage
\appendix
\onecolumn
\section{Experimental Setup}\label{sec:appendix_Experimental}

This section provides a detailed overview of the experimental setup used in our study, including the pruning techniques, datasets, sparsity ranges, and computational environment.

We employed a range of established one-shot pruning techniques, which perform pruning in a single step, followed by Hessian-based updates of the remaining weights and reduce the impact on loss after pruning.
Specifically, we considered WoodFisher~\cite{WoodFisher}, CBS~\cite{CBS}, CHITA~\cite{CHITA}, and Matrix-Free Approximate Curvature (M-FAC)~\cite{mfac}. Performance metrics for these methods were sourced from existing literature~\cite{CBS,CHITA}, with results averaged over three independent runs.

\textbf{Application of \REFLOW{}:}  
In this work, \REFLOW{} is applied to networks pruned using \emph{magnitude pruning}. After pruning, Batch Normalization (BN) running statistics are recalibrated using a forward pass over a limited number of training samples. 

\textbf{Hyperparameters:}  
For \REFLOW{}, we used 50 training batches to recalibrate the running BN statistics, with a batch size of 128 across all experiments.

\textbf{Pre-Trained Networks and Datasets:}  
To ensure comparability with prior studies~\cite{CBS,CHITA}, we adopted datasets and model architectures from the same studies. The analysis included three pre-trained networks: ResNet-20~\cite{RESNET} trained on the CIFAR-10 dataset~\cite{CIFAR10}, and MobileNet~\cite{MobileNet} and ResNet-50~\cite{RESNET} trained on the ImageNet dataset~\cite{ImageNet}. 

We extended the analysis to include larger architectures that prior leading one-shot pruning methods~\cite{WoodFisher, CBS} did not explore and are unable to scale to efficiently. Specifically, we evaluated \REFLOW{} on ResNet-101~\cite{RESNET}, ResNet-152~\cite{RESNET}, RegNetX~\cite{radosavovic2020designing}, and ResNeXt-101~\cite{xie2017aggregated}, all trained on the ImageNet dataset.

\textbf{Sparsity Range:}  
We evaluated \REFLOW{} across the following sparsity ranges, consistent with prior works~\cite{CBS,CHITA}:
\begin{itemize}
    \item \textbf{ResNet-20 on CIFAR-10:} Sparsity range of 0.4 to 0.9.
    \item \textbf{MobileNet on ImageNet:} Sparsity range of 0.4 to 0.8.
    \item \textbf{ResNet-50 on ImageNet:} Sparsity range of 0.4 to 0.9.
\end{itemize}

\textbf{Hardware:}  
All experiments were conducted on a computational setup comprising an NVIDIA RTX A6000 GPU with 48GB memory, 10,752 CUDA cores, and 336 Tensor cores capable of 309 TFLOPS peak performance, coupled with an AMD EPYC 7713P 64-Core CPU.

\textbf{Software:}  
The computational environment operated on Ubuntu 20.04.6 LTS (Focal Fossa), utilizing Python version 3.8.10 and PyTorch version 2.1.0.

\section{Ablation Studies}\label{sec:appendix}
In this section, we evaluate the performance of \REFLOW{} through ablation studies. We analyze the impact of the number of training batches (\(N\)), layer-wise BN recalibration, and batch size on accuracy recovery in pruned networks.

\subsection{Effect of the Number of Training Batches on Performance}\label{sec:Ablation_batches}

We analyze the impact of varying the number of training batches (\(N\)) on the performance of \REFLOW{}, focusing on test accuracy. \REFLOW{} is applied to sparse networks after magnitude pruning, recalibrating Batch Normalization (BN) statistics through a forward pass over \(N\) training batches. 

\begin{figure}[t]
\centering
\begin{tikzpicture}
    \begin{semilogxaxis}[
        width=12cm,
        height=0.11\textwidth,
        scale only axis,
        grid=major,
        grid style={dashed,gray!30},
        tick label style={font=\footnotesize},
        label style={font=\footnotesize},
        legend style={font=\footnotesize},
        cycle list name=color list,
        legend cell align=left,
        xlabel={Number of Training Batches $(N)$},
        ylabel={Test Accuracy (\%)},
        xmin=1, xmax=1000,
        ymin=0, ymax=50,
        ytick={0, 10, 20, 30, 40, 50},
        log basis x={10}
    ]

    \addplot[color=blue, mark=*, solid, thick] coordinates {
        (1, 0.136)
        (2, 0.194)
        (5, 2.012)
        (10, 19.148)
        (20, 41.022)
        (30, 43)
        (40, 43)
        (50, 43)
        (75, 43)
        (100, 43)
        (200, 43)
        (500, 43)
        (1000, 43)
    };

    \end{semilogxaxis}
\end{tikzpicture}
\caption{Test accuracy of MobileNet at 80\% sparsity using \REFLOW{} for different numbers of training batches (\(N\)). Accuracy improves significantly for \(N \leq 20\), saturates around \(N = 50\), and stabilizes for larger \(N\).}\label{fig:mobnet_accuracy_batches}
\end{figure}
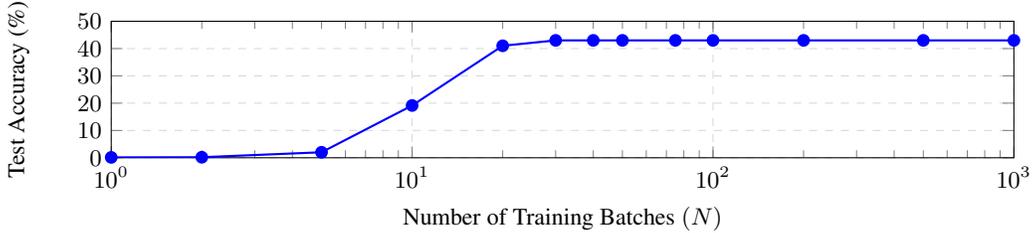

Figure~\ref{fig:mobnet_accuracy_batches} shows the relationship between \(N\) and test accuracy for MobileNet at 80\% sparsity. Accuracy improves significantly for small values of \(N\), saturating around \(N = 50\). Using \(N = 50\) training batches with a batch size of 128 corresponds to only 6,400 images—less than 0.5\% of the 1.28 million training samples in ImageNet.

In contrast, leading impact-based pruning methods such as WoodFisher \cite{WoodFisher} and CBS \cite{CBS} require 960,000 training samples for gradient computation, while CHITA \cite{CHITA} requires 16,000 samples. \REFLOW{} achieves comparable performance using just 6,400 samples without any gradient computation, relying solely on forward passes to update BN statistics. This minimal data requirement enables \REFLOW{} to operate in scenarios where access to the full training dataset is limited, such as privacy-preserving applications or resource-constrained environments, where re-training is infeasible.

\subsection{Impact of Layer-wise Recovery on Performance}\label{sec:Ablation_layerwise}

To gain deeper insights into the recovery of test accuracy in sparse networks, we analyzed the contribution of individual Batch Normalization (BN) layers by recalibrating them sequentially. Specifically, the recalibration was performed one layer at a time, measuring the cumulative improvement in test accuracy after recalibrating each BN layer. This process was conducted in two directions: from the first BN layer to the last (forward direction) and from the last BN layer to the first (backward direction).

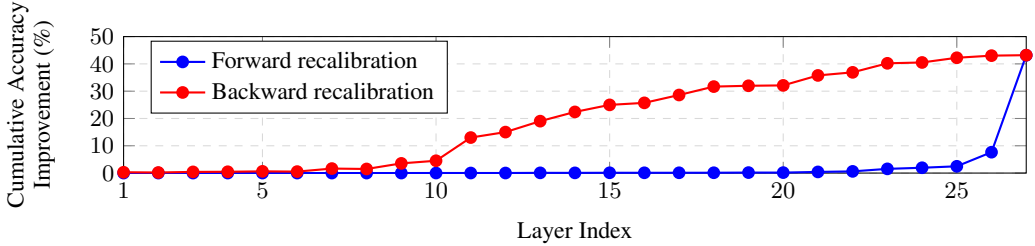
\begin{figure}[t]
\centering
\begin{tikzpicture}
    \begin{axis}[
        width=12cm,
        height=0.11\textwidth,
        scale only axis,
        grid=major,
        grid style={dashed,gray!30},
        tick label style={font=\footnotesize},
        label style={font=\footnotesize},
        legend style={font=\footnotesize},
        cycle list name=color list,
        legend cell align=left,
        xlabel={Layer Index},
        ylabel style={align=center},
        ylabel={Cumulative Accuracy \\Improvement (\%)},
        xmin=1, xmax=27,
        ymin=0, ymax=50,
        xtick={1, 5, 10, 15, 20, 25},
        ytick={0, 10, 20, 30, 40, 50},
        legend pos=north west
    ]

    \addplot[color=blue, mark=*, solid, thick] coordinates {
        (1, 0.0)
        (2, -0.0)
        (3, -0.0)
        (4, 0.0)
        (5, 0.0)
        (6, -0.0)
        (7, -0.0)
        (8, -0.0)
        (9, 0.01)
        (10, 0.02)
        (11, 0.02)
        (12, 0.02)
        (13, 0.08)
        (14, 0.07)
        (15, 0.11)
        (16, 0.1)
        (17, 0.11)
        (18, 0.12)
        (19, 0.18)
        (20, 0.17)
        (21, 0.42)
        (22, 0.62)
        (23, 1.54)
        (24, 1.95)
        (25, 2.48)
        (26, 7.65)
        (27, 43.15)
    };

    \addplot[color=red, mark=*, solid, thick] coordinates {
        (1, 0.3)
        (2, 0.21)
        (3, 0.43)
        (4, 0.48)
        (5, 0.63)
        (6, 0.54)
        (7, 1.65)
        (8, 1.5)
        (9, 3.55)
        (10, 4.51)
        (11, 13.01)
        (12, 15.01)
        (13, 19.02)
        (14, 22.4)
        (15, 24.98)
        (16, 25.72)
        (17, 28.6)
        (18, 31.68)
        (19, 31.98)
        (20, 32.12)
        (21, 35.77)
        (22, 36.9)
        (23, 40.21)
        (24, 40.52)
        (25, 42.24)
        (26, 43.02)
        (27, 43.19)
    };

    \legend{Forward recalibration, Backward recalibration}
    \end{axis}
\end{tikzpicture}
\caption{Cumulative accuracy improvement (\%) for MobileNet at 80\% sparsity after one-shot magnitude pruning. Forward recalibration progresses from the first BN layer to the last, while backward recalibration starts from the last BN layer. Backward recalibration achieves significant improvements earlier than forward recalibration, reflecting the higher sensitivity of deeper layers to pruning-induced changes.}\label{fig:layerwise_recovery}
\end{figure}

Figure~\ref{fig:layerwise_recovery} presents the cumulative effect of BN recalibration on test accuracy for MobileNet at 80\% sparsity after one-shot pruning. In the forward direction, recalibrating early BN layers contributes minimally to accuracy recovery, with notable improvements only emerging as deeper layers are recalibrated. This pattern suggests that the shallower layers are less sensitive to changes in their BN statistics, whereas deeper layers play a more critical role in preserving network performance. Conversely, in the backward direction, recalibrating late BN layers produces substantial accuracy gains early on, with diminishing returns as earlier layers are recalibrated. These observations indicate that later layers are disproportionately impacted by pruning-induced changes, reflecting their higher sensitivity.

This behavior aligns with the phenomenon of \emph{signal collapse}, where the variance of activations diminishes significantly in deeper layers of the pruned network. As described in Equation~\ref{eq:signal_collapse_definition}, the variance ratio between pruned and original activations approaches zero in the final layers, leading to near-constant activations. This results in indistinguishable representations, which propagate to the output, causing uniform or incorrect predictions. The pronounced recovery observed when recalibrating the last layers supports this theoretical insight: correcting the BN statistics in these layers mitigates signal collapse, restoring the discriminative power of the network's activations.

\subsection{Effect of Batch Size on Performance}\label{sec:Ablation_batch}

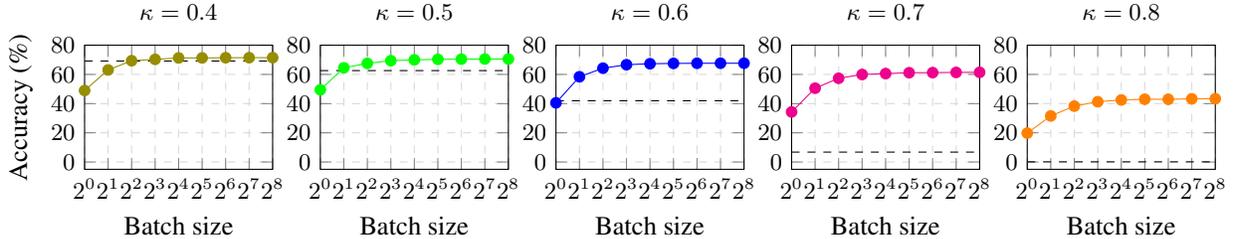
\begin{figure}[t]
\centering
\begin{tikzpicture}
    \begin{groupplot}[
        group style={
            group size=5 by 1,
            vertical sep=35pt,
            horizontal sep=18pt
        },
        width=2.5cm,
        height=0.1\textwidth,
        scale only axis,
        grid=major,
        grid style={dashed,gray!30},
        tick label style={font=\footnotesize},
        label style={font=\footnotesize},
        legend style={font=\footnotesize},
        cycle list name=color list,
        legend cell align=left,
        xmode=log, 
        log basis x={2}, 
        xmin=1, xmax=256,
        ymin=-5, ymax=80,
        xtick={1,2,4,8,16,32,64,128,256},
        xlabel near ticks,
        ylabel near ticks,
    ]
    
    \nextgroupplot[title={$\kappa = 0.4$}, ylabel={Accuracy (\%)}, xlabel={Batch size}, title style={font=\footnotesize}]
    \addplot[color=olive, mark=*] coordinates {
    (1.0, 48.97) (2.0, 63.05) (4.0, 69.38) (8.0, 70.30) (16.0, 71.28) (32.0, 71.29) (64.0, 71.42) (128.0, 71.46) (256.0, 71.48)
    };
    \addplot[color=black, dashed] coordinates {
    (1.0, 69.16) (256.0, 69.16)
    };

    \nextgroupplot[title={$\kappa = 0.5$}, xlabel={Batch size}, title style={font=\footnotesize}]
    \addplot[color=green, mark=*] coordinates {
    (1.0, 49.45) (2.0, 64.52) (4.0, 67.57) (8.0, 69.40) (16.0, 69.97) (32.0, 70.31) (64.0, 70.45) (128.0, 70.53) (256.0, 70.55)
    };
    \addplot[color=black, dashed] coordinates {
    (1.0, 62.61) (256.0, 62.61)
    };

    \nextgroupplot[title={$\kappa = 0.6$}, xlabel={Batch size}, title style={font=\footnotesize}]
    \addplot[color=blue, mark=*] coordinates {
    (1.0, 40.52) (2.0, 58.38) (4.0, 64.25) (8.0, 66.64) (16.0, 67.31) (32.0, 67.54) (64.0, 67.63) (128.0, 67.66) (256.0, 67.66)
    };
    \addplot[color=black, dashed] coordinates {
    (1.0, 41.94) (256.0, 41.94)
    };

    \nextgroupplot[title={$\kappa = 0.7$}, xlabel={Batch size}, title style={font=\footnotesize}]
    \addplot[color=magenta, mark=*] coordinates {
    (1.0, 34.30) (2.0, 50.54) (4.0, 57.34) (8.0, 59.93) (16.0, 60.55) (32.0, 61.14) (64.0, 61.22) (128.0, 61.36) (256.0, 61.43)
    };
    \addplot[color=black, dashed] coordinates {
    (1.0, 6.78) (256.0, 6.78)
    };

    \nextgroupplot[title={$\kappa = 0.8$}, xlabel={Batch size}, title style={font=\footnotesize}]
    \addplot[color=orange, mark=*] coordinates {
    (1.0, 19.80) (2.0, 31.55) (4.0, 38.26) (8.0, 41.28) (16.0, 42.50) (32.0, 42.99) (64.0, 42.95) (128.0, 43.27) (256.0, 43.33)
    };
    \addplot[color=black, dashed] coordinates {
    (1.0, 0.11) (256.0, 0.11)
    };

    \end{groupplot}
\end{tikzpicture}
\caption{Test accuracy of MobileNet at different sparsity levels (\(\kappa\)) and varying batch sizes on ImageNet using \REFLOW{}. Dashed lines represent the baseline accuracy for Magnitude Pruning (MP) without \REFLOW{}.}
\label{fig:batch_acc}
\end{figure}

Here, we investigate the influence of varying batch sizes on the test accuracy of \REFLOW{} for different target sparsity levels (\(\kappa\)) as shown in Figure~\ref{fig:batch_acc}.

At lower sparsity levels (\(\kappa = 0.4\) and \(\kappa = 0.5\)), using smaller batch sizes for \REFLOW{} results in a drop in accuracy below the baseline performance of Magnitude Pruning (MP). This indicates that insufficient recalibration data can negatively impact performance in less sparse networks. However, increasing the batch size leads to a noticeable improvement in accuracy, with \REFLOW{} surpassing MP at moderate and large batch sizes. These results demonstrate that networks with lower sparsity still benefit from recalibration when sufficient batch statistics are available.

For intermediate sparsity (\(\kappa = 0.6\)), the impact of batch size is more pronounced. Accuracy improves consistently with larger batch sizes, significantly outperforming MP even at smaller batch sizes. Saturation occurs at moderate batch sizes, highlighting the increased dependency on recalibration as network sparsity increases.

At higher sparsity levels (\(\kappa = 0.7\) and \(\kappa = 0.8\)), larger batch sizes are critical for achieving substantial gains over MP. Accuracy improves steadily with batch size, with saturation occurring at higher batch sizes compared to lower sparsity levels. These results highlight the importance of recalibration in mitigating the performance degradation caused by high sparsity. The dashed lines in Figure~\ref{fig:batch_acc} provide a reference to the baseline MP performance, underscoring the effectiveness of \REFLOW{} in recovering accuracy, particularly for highly sparse networks.

\section{Analyzing Pruning Similarity Using Hamming Distance}\label{appendix:pruning_similarity}
To further understand the limited role of weight selection, we analyze the  \textit{Normalized Hamming Distance} between pruning masks produced by MP, CHITA, and random pruning. CHITA is used as the representative state-of-the-art (SOTA) IP method.

The \textit{Hamming Distance} between two masks \( m^{(A)} \) and \( m^{(B)} \) is defined as:
\[
H(m^{(A)}, m^{(B)}) = \sum_{i=1}^d \mathbb{I}\left(m^{(A)}_i \neq m^{(B)}_i\right),
\]
where \( \mathbb{I}(\cdot) \) is the indicator function, \( d \) is the total number of parameters, and \( m_i = 1 \) indicates that parameter \( i \) is retained. The \textit{Normalized Hamming Distance}, which measures the fraction of differing pruning decisions between two masks, is defined as:  
\[
H_{\text{norm}}(m^{(A)}, m^{(B)}) = \frac{H(m^{(A)}, m^{(B)})}{d}.
\]
where \( H(m^{(A)}, m^{(B)}) \) is the Hamming Distance, and \( d \) is the total number of parameters.

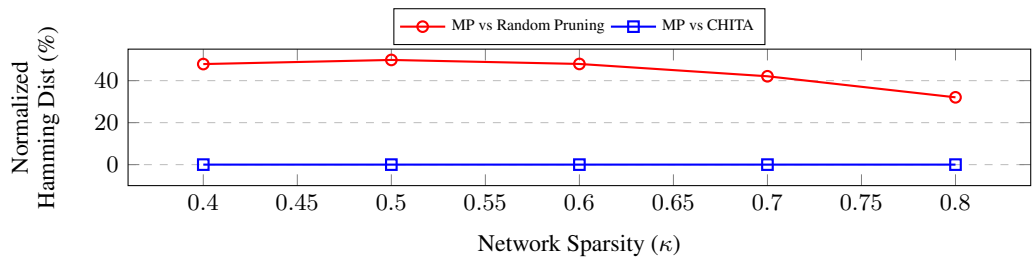
\begin{figure}[h]
\centering
\begin{tikzpicture}
\begin{axis}[
    width=12cm,
    height=0.11\textwidth,
    scale only axis,
    xlabel={Network Sparsity (\(\kappa\))},
    ylabel={Normalized \\ Hamming Dist (\%)},
    ylabel style={align=center},
    ymin=-10, ymax=55,
    ymajorgrids=true,
    grid style=dashed,
    legend pos=south east,
    legend style={
        at={(0.5,+1.3)},
        anchor=north,
        font=\tiny,
        cells={anchor=west},
        inner sep=2pt,
        legend columns=4,
    },
    ymajorgrids=true,
    grid style=dashed,
    tick label style={font=\footnotesize},
    label style={font=\footnotesize},
    legend cell align=left,
    mark options={scale=1},
    cycle list name=color list
]

\addplot[color=red, mark=o, solid, thick] coordinates {
    (0.4, 47.8897)
    (0.5, 49.8520)
    (0.6, 47.9436)
    (0.7, 42.0788)
    (0.8, 32.0510)
};
\addlegendentry{MP vs Random Pruning}

\addplot[color=blue, mark=square, solid, thick] coordinates {
    (0.4, 0.0018)
    (0.5, 0.0018)
    (0.6, 0.0022)
    (0.7, 0.0030)
    (0.8, 0.0030)
};
\addlegendentry{MP vs CHITA}

\end{axis}
\end{tikzpicture}
\caption{Normalized Hamming Distance (\%) between pruning masks for Magnitude Pruning (MP) vs Random pruning and MP vs CHITA across sparsity levels. MP and CHITA have negligible variation, while MP and Random pruning show significant differences.}
\label{fig:normalized_hamming}
\end{figure}

Figure~\ref{fig:normalized_hamming} shows that the Normalized Hamming Distance between MP and CHITA is negligible, indicating close similarity in their pruning decisions compared to the significant variation with random pruning. For ResNet-20 on CIFAR-10, it is \(0.0018\%\). For MobileNet on ImageNet, it is \(0.0095\%\). These results show that magnitude-based and IP-selection methods make nearly identical pruning decisions, supporting the conclusion that the choice of weight selection (MP or IP-selection) has minimal influence on pruning performance.

\label{appendix:pruning_similarity}

\end{document}